\documentclass[sigconf]{acmart}
\usepackage{booktabs} 
\usepackage{amsmath, amsfonts}
\usepackage{amssymb}
\usepackage{dsfont}
\usepackage{bm}
\usepackage[inline]{enumitem}
\usepackage{cleveref}
\usepackage{multirow}
\usepackage{graphicx}
\usepackage{algorithm}
\usepackage[noend]{algpseudocode}
\usepackage{mathtools,tikz,caption}
\DeclareRobustCommand\sampleline[1]{%
  \tikz\draw[#1] (0,0) (0,\the\dimexpr\fontdimen22\textfont2\relax)
  -- (2em,\the\dimexpr\fontdimen22\textfont2\relax);%
}

\makeatletter
\def\BState{\State\hskip-\ALG@thistlm}
\makeatother

\def\etal{\emph{et al.}}
\def\ie{\emph{i.e.}}
\def\eg{\emph{e.g.}}

\definecolor{forestgreen}{rgb}{0.13, 0.55, 0.13}
\definecolor{cornellred}{rgb}{0.7, 0.11, 0.11}
\definecolor{pastelorange}{rgb}{1.0, 0.7, 0.28}
\newcommand{\norm}[1]{\left\lVert#1\right\rVert}
\DeclareMathOperator{\tr}{tr}

\DeclareMathOperator{\argminop}{argmin}

\newcommand{\logdet}{\log\det}
\newcommand{\fronorm}[1]{\left\lVert#1\right\rVert_F^2}
\newcommand{\cov}{\Sigma}

\newcommand{\R}{\mathbb{R}}

\def\Ecoli{\emph{Escherichia coli}}
\def\ecoli{\emph{E. coli}}
\def\ltgl{LTGL}
\graphicspath{{./images_type1/}}
\setcopyright{acmcopyright}
\copyrightyear{2018}
\acmYear{2018}
\setcopyright{acmcopyright}
\acmConference[KDD '18]{The 24th ACM SIGKDD International Conference on Knowledge Discovery \& Data Mining}{August 19--23, 2018}{London, United Kingdom}
\acmBooktitle{KDD '18: The 24th ACM SIGKDD International Conference on Knowledge Discovery \& Data Mining, August 19--23, 2018, London, United Kingdom}
\acmPrice{15.00}
\acmDOI{10.1145/3219819.3220121}
\acmISBN{978-1-4503-5552-0/18/08}

\newcommand{\showDOI}[1]{\unskip}
\newcommand{\shownote}[1]{\unskip}
\newcommand{\showURL}[1]{\unskip}
\newcommand{\showauthor}[1]{\unskip}

\begin{document}

\fancyhead{}
\title{Latent Variable Time-varying Network Inference}

\author{Federico Tomasi}
\authornote{These authors equally contributed to the paper.}
\orcid{0000-0002-8718-3844}
\affiliation{%
  \institution{Universit\`a degli Studi di Genova}
  \streetaddress{Via Dodecaneso 35}
  \city{16146 Genova}
  \state{Italy}
  \postcode{16146}
}
\email{federico.tomasi@dibris.unige.it}

\author{Veronica Tozzo}
\authornotemark[1]
\orcid{0000-0001-8538-9198}
\affiliation{%
  \institution{Universit\`a degli Studi di Genova}
  \streetaddress{Via Dodecaneso 35}
  \city{16146 Genova}
  \state{Italy}
  \postcode{16146}
}
\email{veronica.tozzo@dibris.unige.it}

\author{Saverio Salzo}
\affiliation{%
  \institution{Istituto Italiano di Tecnologia}
  \streetaddress{Via Morego 30}
  \city{16163 Genova}
  \state{Italy}
  \postcode{16163}
}
\email{saverio.salzo@iit.it}

\author{Alessandro Verri}
\affiliation{%
  \institution{Universit\`a degli Studi di Genova}
  \streetaddress{Via Dodecaneso 35}
  \city{16146 Genova}
  \state{Italy}
  \postcode{16146}
}
\email{alessandro.verri@unige.it}

\renewcommand{\shortauthors}{Tomasi, Tozzo \etal}

\begin{abstract}
In many applications of finance, biology and sociology, complex systems involve entities interacting with each other.
These processes have the peculiarity of evolving over time and of comprising latent factors, which influence the system without being explicitly measured.
In this work we present
\emph{latent variable time-varying graphical lasso} (\ltgl{}),
a method for multivariate time-series graphical modelling that considers the influence of hidden or unmeasurable factors.
The estimation of the contribution of the latent factors is embedded in the model which produces both sparse and low-rank components for each time point.
In particular, the first component represents the connectivity structure of observable variables of the system, while the second represents the influence of hidden factors, assumed to be few with respect to the observed variables.
Our model includes temporal consistency on both components, providing an accurate evolutionary pattern of the system.
We derive a tractable optimisation algorithm based on alternating direction method of multipliers, and develop a scalable and efficient implementation which exploits proximity operators in closed form.
\ltgl\ is extensively validated on synthetic data, achieving optimal performance in terms of accuracy, structure learning and scalability with respect to ground truth and state-of-the-art methods for graphical inference. 
We conclude with the application of \ltgl\  to real case studies, from biology and finance, to illustrate how our method can be successfully employed to gain insights on multivariate time-series data.
\end{abstract}

%
%



\maketitle

\section{Introduction}
The problem of understanding complex systems arises in many diverse contexts, such as financial markets \cite{liu2012transelliptical, orchard2013bayesian}, social networks \cite{farasat2015probabilistic} and biology \cite{huang2016inference, hecker2009gene}.
In such contexts, the goal is to analyse the system in order to retrieve information on how the components behave. This requires accurate and interpretable mathematical models whose parameters, in practice, need to be estimated from observations.

Mathematically, a system can be modelled as a network of interactions (edges) between its entities (nodes).
%
%
However, the underlying structure of the variables within the system is usually not known \textit{a priori}. Nevertheless, observations of the system (\ie, data) incorporate information on the interactions between variables, since they provide measurements of such variables acting in the system.

The problem of inferring a network of variable interactions from data is known as \emph{network inference} or \emph{graphical model selection} \cite{lauritzen1996graphical, friedman2008sparse}.
During the last years, the graphical modelling problem has received much attention, particularly for the availability of an always increasing number of samples that are required for a reliable network inference.
Nonetheless, structure estimation of complex systems remains challenging for many reasons.
In this work, we want to tackle two aspects:
\begin{enumerate*}[label=\emph{(\roman*)}]
 \item\label{item-latent} the presence of global hidden (or \emph{latent}) factors, and
 \item\label{item-time} the dynamic of systems that evolve in time.
\end{enumerate*}
We argue that the inference of a dynamical network encoding a complex system requires a specific attention to both aspects to result in a more realistic representation.
In particular, a system may be affected by (latent) factors not encoded in the model. Such factors, acting in the system, influence how the observable entities behave and, hence, how they are connected with each other \cite{choi2010gaussian}.
The consideration of hidden and unmeasured variables during the inference process emerges as crucial to avoid misrepresenting real-world data \cite{meng2014learning}.

At the same time, a complex system depends on a temporal component, which drives variable interactions to evolve consistently during its extent.
This means that the structure can either change or remain stable according to the nature of the system itself.
%
Hence, the understanding of a complex system is bound to the observation of its evolution.
This is particularly evident in some applications, such as biology, where the interest could be to understand the response of the system to perturbation \cite{molinelli2013perturbation}. 
%
\paragraph{\bf Related work}
Latent variable models have been widely studied in literature,
and shown to outperform graphical models that only consider observable variables \cite{chandrasekaran2010latent, choi2011learning, yuan2012}. At the same time, a set of methods were designed to study the temporal component through the inference of a dynamical network that incorporates prior knowledge on the behaviour of the system \cite{bianco2016successful, hallac2017network}.
Time-series with latent variables are considered to obtain a single graph which represents the global system \cite{jalali2011learning, anandkumar2013learning}.
However, to the best of our knowledge, state-of-the-art methods for regularised network inference do not consider simultaneously time and latent variables in the inference of multiple connected networks.
\paragraph{\bf Contribution}
In this work we propose \emph{latent variable time-varying graphical lasso} (\ltgl{}), a model for dynamical network inference where the structure is influenced by latent factors.
This can be seen as an attempt to generalise both dynamical and latent variable network inference under a single unified framework.
In particular, starting from a set of observations of a system at different time points, \ltgl\ infers an interaction network of the observed variables under the influence of latent factors, while taking in consideration the temporal evolution of the system.
%
The empirical interaction network is decomposed into the true underlying structure of the network and the contribution of latent factors,
under the assumption that both observable variables and latent factors interdependence follow a temporal non-random behaviour.
%
%
%
For this reason, the model allows to include prior knowledge on the evolutionary pattern of the system.
The imposition of such prior knowledge benefits inference and subsequent analysis of the network,
accentuating precise dynamical patterns.
This is particularly important when the number of samples is low compared to the number of observed and latent variables in the system.
In fact,
the inference of the network at particular time points exploits the dependence between consecutive temporal states. Such advantage is achieved by a simultaneous inference of all the dynamical system, that, mathematically, translates into imposing constraints on the network behaviour.
In this work we provide a set of possible constraints that can be applied independently on both observed and latent components, allowing for a wide range of evolutionary patterns.
%

\begin{figure}
\includegraphics[scale=0.5]{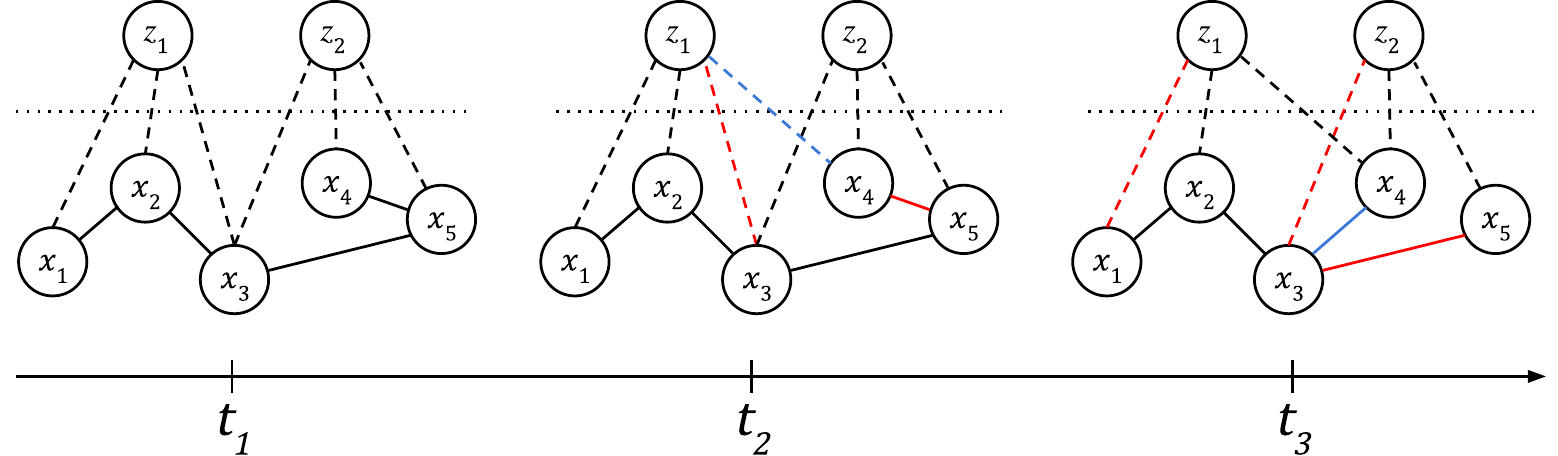}
\vspace*{-.5em}
\caption{\small{A dynamical network with latent factors $z_i$ and observed variables $x_i$.
At each time $t_i$, all of the connections between latent and observed variables (\sampleline{dashed} lines) and connections among observed variables (\sampleline{} lines) may change according to a specific temporal behaviour.}}
\label{fig:example}
\vspace*{-1.5em}
\end{figure}

\Cref{fig:example} provides an example of the theoretical model assumed by \ltgl{}. Here, observed and latent variables ($x_i$ and $z_i$) are connected in a slightly different way at each time.
Note that the observations of the system only involve variables $x_i$, while the hidden factors $z_i$ influence the system without being actually observed.
Hence, when analysing samples which are regulated from a dynamical network with hidden factors, it is infeasible to precisely infer the identity of latent variables, but only an estimation of their number and their effect on the global system can be obtained.

Starting from the theoretical model we derived a minimisation algorithm based on the alternating direction method of multipliers (ADMM) \cite{Boyd2010}.
The algorithm is divided into independent steps using proximal operators, which can be solved by closed-form solutions favouring a fast and scalable computation \cite{Ma2013, Danaher2014, hallac2017network}.
%
We also provide the related implementation in a Python framework, based on the use of highly optimised low-level libraries for numerical computation.
%
%
Experiments on synthetic data show \ltgl\ to achieve optimal performance in relation to ground truth and to state-of-the-art methods for graphical modelling, in terms of accuracy, structure learning and scalability.
Moreover, we show the computational efficiency of \ltgl\ while increasing the number of unknowns of the problem and the model complexity.
We conclude with the application of \ltgl\ to real-world data sets
to illustrate how our method can be successfully employed to gain insights on multivariate time-series data.
In particular, we used biological and financial data sets, to show the use of \ltgl\ in different contexts. In the first case, we analysed \Ecoli\ response to perturbation, correctly identified by our method. In the latter case, we investigated on a financial data set, to show how the contribution of latent factors is relevant for the understanding of the behaviour of the system.

\paragraph{\bf Outline} The paper is organised as follows.
\Cref{sec:preliminaries} includes a background on the reference frameworks for static and dynamical network inference.
\Cref{contribution} contains the theoretical formulation of the problem and the proposed method.
\Cref{mm} describes in details the optimisation algorithm for the minimisation of the functional.
\Cref{synthetic} and \Cref{real} illustrate the use of our method on synthetic and real data, respectively.
Finally, \Cref{sec:conclusions} concludes with a discussion and future directions.

%
%

\section{Preliminaries}\label{sec:preliminaries}
Given a graph  $\mathcal{G} = (\mathcal{V}, \mathcal{E})$, where $\mathcal{V} = \{x_1,\dots,x_d\}$ is a finite set of vertices, and $\mathcal{E} \subseteq \mathcal{V}\times\mathcal{V}$ is a set of edges, a \emph{graphical model} is a multivariate probability distribution on $x_1, \dots, x_d$ variables where the conditional independence between two variables $x_i$ and $x_j$ given all the others is encoded in $\mathcal{G}$ \cite{lauritzen1996graphical}. The two variables $x_i$ and $x_j$ are conditionally independent given the others if $(x_i, x_j) \notin \mathcal{E}$ and $(x_j, x_i) \notin \mathcal{E}$.
In what follows, we consider only undirected Gaussian graphical models (GGMs), where \begin{enumerate*}[label=\emph{(\roman*)}]\item  there is no distinction between an edge $(x_i, x_j)\in\mathcal{E}$ and $(x_j, x_i)$, and
\item variables are jointly distributed according to a multivariate Gaussian distribution $\mathcal{N}(\mu, \cov)$. 
\end{enumerate*}
Without loss of generality, we assume $\mu$ to be zero, 
thus the distribution depends only on the covariance matrix $\cov$ \cite{choi2011learning}.
The inverse covariance matrix $\Theta = \cov^{-1}$, called \emph{precision} matrix, encodes the conditional independence between pairs of variables. In particular, the precision matrix has a zero entry in the position $i,j$ only if $(x_i,x_j) \notin \mathcal{E}$ \cite{lauritzen1996graphical}. 
Hence, we can interpret the precision matrix as the weighted adjacency matrix of $\mathcal{G}$, encoding the dependence between variables.
%
\paragraph{\bf Network inference}\sloppy\relax
Consider a series of samples drawn from a multivariate Gaussian distribution $X  \sim \mathcal{N}(\mathbf{0}, \cov)$, $X\in \mathbb{R}^{n \times d}$.
Network inference aims at recovering the graphical model of the $d$ variables, \ie, the interaction structure $\Theta$, given $n$ observed samples.
The graphical modelling problem has been extensively tackled in literature by estimating the precision matrix $\Theta$ instead of the covariance matrix $\cov$ (as, \eg, \cite{Bien2011}) \cite{yuan2007model, friedman2008sparse, ravikumar2011high}.
%
This has been shown to improve the graphical model inference, particularly for high-dimensional problems \cite{meinshausen2006high}.
In such contexts, the assumption is that a variable is conditionally dependent only on a subset of all the others. Therefore, the estimation of the precision matrix may be guided by a sparse prior, in such a way to restrict the number of possible connections in the network to improve interpretability and noise reduction.
Also, the imposition of a sparse prior on the problem helps with the identifiability of the graph, especially when the available number of samples is low compared to the dimension of the problem.
A model for the inference of $\Theta$ including the sparse prior is the \emph{graphical lasso} \cite{friedman2008sparse}:
\small\begin{equation}\label{eq:graphical-lasso}
\underset{
\begin{subarray}{c}
\Theta
\end{subarray}
}{\operatorname{minimize}}
~- \ell(S, \Theta) + \alpha\|\Theta\|_{od, 1},
\end{equation}\normalsize
where $\ell$ is the Gaussian log-likelihood (up to a constant and scaling factor) defined as
$\ell(S, \Theta) = \log\det(\Theta) - \tr(S\Theta)$ for $\Theta$ positive definite and $S=\frac{1}{n} X^\top X$ is the empirical covariance matrix.
\mbox{$\|\cdot\|_{od, 1}$} is the off-diagonal $\ell_1$-norm, which promotes sparsity in the precision matrix (excluding the diagonal).

\paragraph{\bf Latent variable network inference}
Often, real-world observations do not conform exactly to a sparse GGM. This is due to global hidden factors that influence the system, which introduce spurious dependencies between observed variables \cite{choi2010gaussian, choi2011learning}.
For this reason, GGMs can be extended by introducing latent variables able to represent factors which are not observed in the data.
These latent variables are \emph{not} principal components, since
they do not provide a low-rank approximation of the graphical model.
On the contrary, such factors are added to the model in order to condition the statistics of the observed variables.
%
In particular, we consider both latent and observed variables to have a common domain \cite{choi2011learning}.

Let latent variables be indexed by a set $H$, and observed variables by a set $O$. The precision matrix $\Theta$ of the joint distribution of both latent and observed variables may be partitioned into four blocks: 
\small
\begin{equation*}\label{matrixform}
\Theta = \Sigma^{-1} = \left[\begin{array}{c|ccc}
&\hspace{-.5em}&&\hspace{-.5em}\\[-.6em]
\Theta_{H} &\hspace{-.5em}& \Theta_{HO}&\hspace{-.5em}\\[.5em]
\hline
&\hspace{-.5em}&&\hspace{-.5em}\\
\Theta_{OH} &\hspace{-.5em}&\Theta_{O}&\hspace{-.5em}\\
&\hspace{-.5em}&&\hspace{-.5em}\\
\end{array}\right].
\end{equation*}
\normalsize
Such blocks represent the conditional dependencies among latent variables ($\Theta_{H}$), observed variables ($\Theta_{O}$), between latent and observed ($\Theta_{HO}$), and viceversa ($\Theta_{OH}$).
The \emph{marginal precision matrix} $\Sigma_{O}^{-1}$ of the observed variables is given by the Schur complement w.r.t. the block $\Theta_{H}$ \cite{chandrasekaran2010latent}:
%
\small\begin{equation}\label{schur}
\hat{\Theta}_{O} = \Sigma_{O}^{-1} = \Theta_{O} - \Theta_{OH}\Theta_{H}^{-1}\Theta_{HO}.
\end{equation}\normalsize
$\Theta_{O}$ specifies the precision matrix of the \emph{conditional statistics} of the observed variables given the latent variables, while $\Theta_{OH}\Theta_{H}^{-1}\Theta_{HO}$ is a \emph{summary} of the effect of marginalisation over the latent variables.
Such matrix has a small rank if the number of latent variables is small compared to the observed variables. Note that the rank is an indicator of the number of latent variables $H$ \cite{chandrasekaran2011rank}.

The effect of the marginalisation is scattered over many observed variables, in such a way not to confound it with the true underlying conditional sparse structure of $\Theta_O$  \cite{chandrasekaran2010latent}.
Typically $\hat{\Theta}_{O}$ is not sparse due to the low-rank term, while, with the addition of the latent factors contribution, we can recover the true sparse GGM.
For this reason, the graphical lasso in \Cref{eq:graphical-lasso} has been extended with the \emph{latent variable graphical lasso} that includes the inference of a low-rank term, using the form \eqref{schur}, as follows \cite{chandrasekaran2010latent}:
\small
\begin{equation}\label{eq:latent}
\tilde{\Theta}, \tilde{L} = \underset{
\begin{subarray}{c}
(\bm{\Theta}, \bm{L}) \\
L \succcurlyeq 0
\end{subarray}
}{\operatorname{argmin}}
~- \ell(S, \Theta-L) + \alpha\|\Theta\|_{od, 1} + \tau\|L\|_*.
\end{equation}
\normalsize

%
%
%
%
Here $\tilde{\Theta}$ provides an estimate of $\Theta_{O}$ (precision matrix of the observed variables) while $\tilde{L}$ provides an estimate of $\Theta_{OH}\Theta_{H}^{-1}\Theta_{HO}$
(marginalisation over the latent variables).
Note that $S$ is the empirical covariance matrix computed only on observed variables, since no information on the latent ones is available.

\paragraph{\bf Time-varying network inference}\label{sec:tvgl}
Problems~\eqref{eq:graphical-lasso} and \eqref{eq:latent} aim at recovering the structure of the system at fixed time (\emph{static network inference}).
However, complex systems have temporal dynamics that
regulate their overall functioning \cite{Albert2007, friedman2008sparse}.
Hence, the modelling of such complex systems requires a \emph{dynamical network inference}, where the states of the network are co-dependent.
This naturally leads to the idea of \emph{temporal consistency},
which assumes similarities between consecutive states of the network.
In fact, we can assume that, for sufficiently close time points, a system shows negligible differences.
During the inference of a dynamical network, temporal consistency may translate into
the imposition of
similarities among temporally close networks \cite{gibberd2017multiple}. In particular, graphical lasso with temporal consistency results in \emph{time-varying graphical lasso}  \cite{hallac2017network}:
\small{\begin{align}\label{dynamicallasso}
\begin{split}
\underset{({\Theta}_1, \dots, {\Theta}_T)}{\operatorname{minimize}}\ \  &\sum_{i=1}^T \bigg[- \ell({S}_i, {\Theta}_i) + \alpha\|{\Theta}_i\|_{od,1} \bigg] + \beta\sum_{i=1}^{T-1} \Psi({\Theta}_{i+1} - {\Theta}_{i})
\end{split}
\end{align}}\normalsize
where the inference of a network at a single time point $i$ is guided by the states at adjacent time points.
%
The network is encoded in a sequence of precision matrices $(\Theta_1,\dots,\Theta_T)$ which model the system at each time point $i=1,\dots,T$.
The type of similarity imposed to consecutive time points and its strength are specified by the penalty function $\Psi$ and the parameter $\beta$, respectively.
Options when choosing $\Psi$ include the following \cite{hallac2017network}:\\[.3em]
{\it\textbullet{} Lasso penalty ($\ell_1$) - $\Psi=\sum_{ij}|\cdot|$}.\\
Encourages few edges to change between subsequent time points, while the rest of the structure remains the same \cite{Danaher2014}.\\[.3em]
{\it\textbullet{} Group lasso penalty - $\Psi=\sum_{j}\|\cdot_j\|_2$}.\\
Encourages the graph to restructure at some time points and to stay stable in others \cite{hallac2015network, gibberd2017multiple}.\\[.3em]
{\it\textbullet{} Laplacian penalty - $\Psi=\sum_{ij}(\cdot_{ij})^2$}.\\
Encourages smooth transitions over time, for slow changes of the global structure \cite{weinberger2007graph}.\\[.3em]
{\it\textbullet{} Max norm penalty ($\ell_\infty$) - $\Psi=\sum_{j}(\max_i|\cdot_{ij}|)$}.\\
Encourages a block of nodes to change their structure with no additional penalty with respect to the change of a single edge among such nodes.
In fact, $\ell_\infty$ norm is influenced only from the most changing element for each row.\\[.3em]
{\it\textbullet{} Row-column overlap penalty - $\Psi=\min_{V:A=V+V^\top}\sum_{j}\|V_j\|_p$}.\\
Encourages a major change of the network at a specific time, while in the rest the network is enforced to remain constant. The choice of $p=2$ causes the penalty to be node-based, \ie, the penalty allows for a perturbation of only some nodes
\cite{mohan2012structured}.
\\[.3em]
Dependently from prior assumptions on the problem, one may choose the most appropriate penalty for the data at hand.

\section{Model formulation}\label{contribution}
In this work we propose a new statistical model for the inference of networks that change consistently in time under the influence of latent factors.
We call such model \emph{latent variable time-varying graphical lasso} (\ltgl{}).
\ltgl\ infers the dynamical network of complex systems by decomposing the problem into two parts --- similarly to what has been done in \cite{chandrasekaran2010latent} for static network inference.
We consider two components of the dynamical network: a true underlying structure on the observed variables and the contribution of latent factors.
This allows to factor out the contribution of hidden variables, favouring a reliable modelling of the dynamical system.
The novelty of our method is the simultaneous inference of a dynamical network with latent factors that exploits the imposition of behavioural consistency on both observed variables interactions and latent influence through the use of penalisation terms.
This allows for an easier interpretation of the evolution of the dynamical system while, at the same time, improving its graphical modelling.
The two separate (while closely related) components at each time point are obtained by
integrating the network inference with the
information coming from temporally different states of the network.

Formally, let $X_i\in \mathbb{R}^{n_i \times d}$, for $i={1, \dots, T}$, be a set of observations measured at $T$ different time points composed by $n_i$ samples of $d$ observed variables.
(Note that, for each time point $i$, samples are assumed to be drawn from the probability distribution on the observed variables conditioned on the latent ones.)
Let $S_i = \frac{1}{n_i}X_i^\top X_i$ be the empirical covariance matrix at time $i$.
The goal is to retrieve a set of sparse matrices $\mathbf{\Theta} = (\Theta_1, \dots, \Theta_T)$ and a set of low-rank matrices $\mathbf{L} = (L_1, \dots, L_T)$ such that, at each time point $i$, ${\Theta}_i$ encodes the conditional independences between the observed variables, while ${L}_i$ provides the summary of marginalisation over latent variables on the observed ones.

%
Consider \Cref{eq:latent} at a specific time $i$. Here we want to impose continuity between the structure and the hidden variables contribution in time, therefore we enforce the difference between consecutive graphs
to abide certain constraints by adding two penalisation terms.
Our \ltgl\ model takes the following form:
\small\begin{align}\label{latent-time}
\begin{split}
\underset{
\begin{subarray}{c}
(\bm{\Theta}, \bm{L}) \\
L_i \succcurlyeq 0
\end{subarray}
}{\text{minimize}}\ \ & \sum_{i=1}^T \bigg[- \ell({S}_i, {\Theta}_i- {L}_i) + \alpha\|{\Theta}_i\|_{od, 1} + \tau\| {L}_i\|_* \bigg] \\[-.5em]
& + \beta\sum_{i=1}^{T-1} \Psi({\Theta}_{i+1} - {\Theta}_{i}) + \eta\sum_{i=1}^{T-1} \Phi({L}_{i+1} - {L}_{i}),
\end{split}
\vspace*{-2em}
\end{align}\normalsize
where $\Psi$ and $\Phi$ are both penalty functions that force the structure of the network to change over time according to a certain behaviour, by acting on $\Theta$ and $L$, respectively.
Temporal consistency of both the structure of the network and latent factors contribution is guaranteed by the use of such penalty functions, which benefits the network inference in particular in presence of few available observations of the system.
Possible choices for $\Psi$ and $\Phi$ are listed in \Cref{sec:tvgl}. Their choice is arbitrary and it is based on the prior knowledge on the respective components evolution in the system.
Also, note that $\Psi$ and $\Phi$ are independent, which allows \ltgl\ to model a wide range of dynamical behaviours of complex systems.
%
%
%

\section{Minimisation method}\label{mm}
Problem \eqref{latent-time} is convex, provided that the penalty functions $\Psi$ and $\Phi$ are convex, and it is coercive because of the regularisers. Thus, Problem~\eqref{latent-time} admits solutions.
Nonetheless, its optimisation is challenging in practice due to the high number of unknown matrices involved ($2T$,  for a total of $2T \frac{d(d+1)}{2}$ unknowns of the problem).
A suitable method for the minimisation is ADMM \cite{Boyd2010}. It allows to decouple the variables obtaining a separable minimisation problem which can be efficiently solved in parallel.
The sub-problems
exploit
proximal operators which are (mostly) solvable in closed-form, leading to a simple iterative algorithm.

In order to decouple the involved matrices, we define three dual variables $\bm{R}$, $\bm{Z} = (\bm{Z}_1, \bm{Z}_2)$ and  $\bm{W} = (\bm{W}_1, \bm{W}_2)$ and two projections:
\small\begin{align*}
P_1\colon (\R^{d\times d})^{T} &\to (\R^{d\times d})^{T-1} & P_2\colon (\R^{d\times d})^{T}&\to (\R^{d\times d})^{T-1}\\
\bm{A}&\mapsto (A_1, \dots, A_{T-1}) & \bm{A}&\mapsto (A_2, \dots, A_{T})
\end{align*}\normalsize
Problem \eqref{latent-time} becomes:
\small\begin{align}\label{latent-time-admm}
\begin{split}
\left.\begin{array}{l}
\underset{ \begin{subarray}{c} (\bm{\Theta}, \bm{L}, \bm{R},\bm{Z}, \bm{W}) \\ \bm{L}_i \succcurlyeq 0
\end{subarray}}{\text{minimize}} \begin{aligned}[t]
& \sum_{i=1}^T \bigg[- \ell({S}_i, R_i) + \alpha\|{\Theta}_i\|_{od, 1} + \tau\| {L}_i\|_* \bigg]\vspace{0.2cm} \\
& + \beta\sum_{i=1}^{T-1} \Psi(Z_{2,i} - Z_{1,i} )+ \eta\sum_{i=1}^{T-1} \Phi(W_{2,i} - W_{1,i})\end{aligned} \vspace{0.2cm}\\
\text{s.t.}~~~ \bm{R} = \bm{\Theta} - \bm{L},~~ \bm{Z}_1 = P_1\bm{\Theta},
~~\bm{Z}_2  = P_2\bm{\Theta},
~~\bm{W}_1 = P_1\bm{L},
~~\bm{W}_2  = P_2\bm{L}.
\end{array}\right\}
\end{split}
\end{align}\normalsize
The corresponding augmented Lagrangian is as follows:
\small
\begin{align}\label{lagrangian}
\begin{split}
&\!\!\!\mathcal{L}_\rho(\bm{\Theta}, \bm{L}, \mathbf{R},  \mathbf{Z}, \mathbf{W}, \mathbf{U}) \\
&\!\!\!=\!\sum_{i=1}^{T}
\begin{aligned}[t]
 & \bigg[- \ell({S}_i, R_i) +
 \alpha\|{\Theta}_i\|_{od, 1} + \tau\| {L}_i\|_* + \mathbb{I}(L \succcurlyeq 0) \bigg]
\end{aligned}\\
&\!\!\!+\!\beta\!\sum_{i=1}^{T-1} \Psi(Z_{2,i} - Z_{1,i} )+ \eta \sum_{i=1}^{T-1} \Phi(W_{2,i} - W_{1,i})\\
&\!\!\!+\!\frac{\rho}{2}\!\sum_{i=1}^{T} \!\bigg[\|R_i\! - \! \Theta_i \! + \! L_i  \! +  \!U_{0,i}\|^2 - \|U_{0,i}\|^2\bigg]\\
&\!\!\!+\!\frac{\rho}{2}\!\sum_{i=1}^{T-1} \!\bigg[\|\Theta_i \!-\! Z_{1,i} \! + \! U_{1,i}\|^2\! - \!\|U_{1,i}\|^2 \! + \! \|\Theta_{i+1} \! -\! Z_{2,i} \! + \! U_{2,i}\|^2  \!- \! \|U_{2,i}\|^2\bigg]\\
&\!\!\!+\!\frac{\rho}{2}\!\sum_{i=1}^{T-1} \!\bigg[\|L_i \! -\!  W_{1,i} \! + \! U_{3,i}\|^2 \! - \! \|U_{3,i}\|^2 \! +  \! \|L_{i+1} \! - \! W_{2,i} \! + \! U_{4,i}\|^2 \! - \! \|U_{4,i}\|^2\bigg]
\end{split}
\end{align}\normalsize

where $\bm{U} = (\bm{U}_0, \bm{U}_1, \bm{U}_2, \bm{U}_3, \bm{U}_4)$ are the scaled dual variables.

The ADMM algorithm for Problem \eqref{latent-time-admm} writes down as follows:
\small\begin{align}\label{eq:admm-algorithm}
\begin{split}
\text{\bf for } k = 1, &\dots\\
\mathbf{R}^{k+1} &= \underset{\mathbf{R}}{\argminop}~ \mathcal{L}_\rho( \bm{\Theta}^k, \bm{L}^k, \mathbf{R}, \mathbf{Z}^k,\mathbf{W}^k, \mathbf{U}^k)\\
\bm{\Theta}^{k+1} &= \underset{\mathbf{\Theta}}{\argminop}~ \mathcal{L}_\rho( \bm{\Theta}, \bm{L}^k, \mathbf{R}^{k+1}, \mathbf{Z}^k,\mathbf{W}^k, \mathbf{U}^k)\\
\bm{L}^{k+1} &= \underset{\mathbf{L}}{\argminop}~ \mathcal{L}_\rho( \bm{\Theta}^{k+1}, \bm{L}, \mathbf{R}^{k+1}, \mathbf{Z}^k,\mathbf{W}^k, \mathbf{U}^k)\\
\mathbf{Z}^{k+1} &= \left[\begin{array}{c}\bm{Z}_1^{k+1}\\ \bm{Z}_2^{k+1} \end{array}\right] = \underset{\mathbf{Z}}{\argminop}~ \mathcal{L}_\rho( \bm{\Theta}^{k+1}, \bm{L}^{k+1}, \mathbf{R}^{k+1}, \mathbf{Z},\mathbf{W}^k, \mathbf{U}^k)\\
\mathbf{W}^{k+1} &= \left[\begin{array}{c}\bm{W}_1^{k+1}\\ \bm{W}_2^{k+1} \end{array}\right] = \underset{\mathbf{W}}{\argminop}~ \mathcal{L}_\rho(\bm{\Theta}^{k+1}, \bm{L}^{k+1}, \mathbf{R}^{k+1},  \mathbf{Z}^{k+1},\mathbf{W}, \mathbf{U}^k)\\
\mathbf{U}^{k+1} &=
 \left[\begin{array}{c}\bm{U}_0^{k}\\ \bm{U}_1^{k}\\ \bm{U}_2^{k} \\\bm{U}_3^{k}\\ \bm{U}_4^{k} \end{array}\right] + \left[\begin{array}{c}
\mathbf{R}^{k+1} \!- \bm{\Theta}^{k+1} + \bm{L}^{k+1}\\
P_1\bm{\Theta}^{k+1}  - \bm{Z}_1^{k+1}\\
P_2\bm{\Theta}^{k+1}  - \bm{Z}_2^{k+1}\\
P_1\bm{L}^{k+1}  - \bm{W}_1^{k+1}\\
P_2\bm{L}^{k+1}  - \bm{W}_2^{k+1}
\end{array}\right].
\end{split}
\end{align}\normalsize


\subsection{$\mathbf{R}$ step}
The minimisation problem involving the matrix $R$ in \eqref{eq:admm-algorithm}
can be split into parallel updates, since $\mathcal{L}_\rho(\bm{\Theta}, \bm{L}, \mathbf{R}, \mathbf{Z}, \mathbf{W}, \mathbf{U})$ is separable in the variables $(R_1, \dots, R_T)$. Therefore, each $R_i$ at iteration $k+1$ is \mbox{given by}:
\small
\begin{align}\label{eq:r-step}\begin{split}
R_i^{k+1} = \underset{R} {\argminop}\ &\tr(S_iR) - \logdet(R) +\frac{\rho}{2}\|R - \Theta^k + L^k + U^k_{0,i}\|^2\\
= \underset{R} {\argminop}\ &\tr(S_iR) - \logdet(R) + \frac{\rho}{2}\fronorm{R - A^k_i }\\
= \underset{R} {\argminop}\ &\tr(S_iR) - \logdet(R) + \frac{\rho}{2}\fronorm{R -  \frac{A^k_i + A^{k\top}_i}{2}}
\end{split}\end{align}\normalsize
with $A^k_i = \Theta^k_i - L^k_i - U^k_{0,i}$. Note that the last equality in \eqref{eq:r-step}
follows from the symmetry of $R$ --- which also guarantees the $\logdet$ to be well-defined.
\Cref{eq:r-step} can be explicitly solved. Indeed, Fermat's rule yields:
%
\small\begin{align}\label{riderivative}
\begin{split}
&S_i - \rho \frac{A^k_i + A^{k\top}_i}{2} = R^{-1} - \rho R.
\end{split}
\end{align}\normalsize
Then the solution to \Cref{riderivative} is \cite{Witten2009, Danaher2014,hallac2017network}:
\small
\begin{equation*}
R_i^{k+1} = \frac{1}{2 \rho }V^k\left(- E^k + \sqrt{(E^k)^2 + 4\rho I} \right)V^{k\top}
\end{equation*}\normalsize
where $V^kE^kV^{k\top}$ is the eigenvalue decomposition of $S_i - \rho  \frac{A^k_i + A^{k\top}_i}{2}$.
%

\subsection{$\mathbf{\Theta}$ step}
Likewise the $\mathbf{R}$ step, the update of $\mathbf{\Theta}$ in \eqref{eq:admm-algorithm}
can be done in a parallel fashion, as follows:
\small \begin{align}\label{eq:t-step}\begin{split}
\Theta_i^{k+1} &= \underset{\Theta}{\argminop}\ \begin{aligned}[t]
 &\alpha\|\Theta\|_{od, 1}
+\frac{\rho}{2}\bigg[\fronorm{R_i^k - \Theta + L_i^k + U_{0,i}^k} \\
&+\overline{\delta}_{iT}\fronorm{\Theta - Z_{1,i}^k + U_{1,i}^k}
+\overline{\delta}_{i1}\fronorm{\Theta - Z_{2,i-1}^k + U_{2,i-1}^k}\bigg] \\
\end{aligned}\\
& = \underset{\Theta}{\argminop}\ \alpha\|\Theta\|_{od, 1} + (1 +  \overline{\delta}_{iT}+ \overline{\delta}_{i1}) \frac{\rho}{2}\fronorm{\Theta - B^k_i}
\end{split}\end{align}\normalsize
where $\overline{\delta}_{ij} = 1 - {\delta}_{ij}$, with ${\delta}_{ij}$ Kronecker delta
and
\small $$B^k_i = \frac{L_i^k + R_i^k + U_{0,i}^k + \overline{\delta}_{iT}(Z_{1,i}^k - U_{1,i}^k) + \overline{\delta}_{i1}(Z_{2,i-1}^k -  U_{2,i-1}^k)}{1 +  \overline{\delta}_{iT}+ \overline{\delta}_{i1}}.$$ \normalsize
%
%
%
Problem \eqref{eq:t-step} is solved as:
\small\begin{align*}
\begin{split}
\Theta^{k+1}_{i} &= \text{prox}_{\zeta\|\cdot\|_{od,1}}(B^k_i) = S_{\zeta}(B^k_i)
\end{split}
\end{align*}
\normalsize
with $\zeta=\frac{\alpha}{\rho(1 + \overline{\delta}_{iT} + \overline{\delta}_{i1})}$, and $S_{\zeta}(\cdot)$ element-wise off-diagonal soft-thresholding function.

\subsection{$\mathbf{L}$ step}
The parallel update of $\mathbf{L}$ in \eqref{eq:admm-algorithm} can be written as:
\small
\begin{align}\label{eq:l-step}\begin{split}
L^{k+1}_{i} &= \begin{aligned}[t]&\underset{L}{\argminop}\ \tau \tr(L) + \mathds{I}(L \succcurlyeq 0)
+ \frac{\rho}{2} \bigg[\fronorm{R_i^{k+1} \!- \Theta_{i}^{k+1} + L + U_{i,0}^k} \\
&+\overline{\delta}_{iT}\fronorm{L - W_{1,i}^k + U_{3,i}^k}
+\overline{\delta}_{i1}\fronorm{L - W_{2,i-1}^k + U_{4,i-1}^k}\bigg]
\end{aligned}\\
&= \underset{L}{\argminop}\ \tau  \tr(L) + \mathds{I}(L \succcurlyeq 0) +
(1+ \overline{\delta}_{iT} + \overline{\delta}_{i1})\frac{\rho}{2} \fronorm{L -  C^k_i}\\
&= \underset{L}{\argminop}\ \tau  \tr(L) + \mathds{I}(L \succcurlyeq 0) +
(1+ \overline{\delta}_{iT} + \overline{\delta}_{i1})\frac{\rho}{2} \fronorm{L -  \frac{C^k_i+C^{k\top}_i}{2}}
\end{split}\end{align}\normalsize
where \small $$C^k_i = \frac{\Theta_i^{k+1} \!- R_i^{k+1} \!- U_{0,i}^k + \overline{\delta}_{iT}(W_{1,i} - U_{3,i}) +  \overline{\delta}_{i1}( W_{2,i-1}  - U_{4,i-1})}{1+ \overline{\delta}_{iT} + \overline{\delta}_{i1}}.$$ \normalsize
Note that the last equality in \eqref{eq:l-step} follows from the symmetry of $L$.

Then, the solution to Problem \eqref{eq:l-step} is \cite{Ma2013}:
\begin{equation*}
L_i^{k+1} = V^k\tilde{E}V^{k\top}
\end{equation*}
where
$V^kE^kV^{k\top}$ is the eigenvalue decomposition of $C^k_i$, and
$$\tilde{E}_{jj} = \max\left(E^k_{jj} - \frac{\tau}{\rho(1+ \overline{\delta}_{iT} + \overline{\delta}_{i1})}, ~~0\right).$$

\subsection{$\mathbf{Z}$ and $\mathbf{W}$ step}
%
The dual variables $\mathbf{Z}$ and $\mathbf{W}$ enforce the network to behave in time consistently with the choice of $\Psi$ and $\Phi$, respectively.
$\mathbf{Z}$ is the dual variable of $\mathbf{\Theta}$ while $\mathbf{W}$ is the dual variable of $\mathbf{L}$. For the sake of brevity, we show only the steps regarding the update of $\mathbf{Z}$ --- the update of $\mathbf{W}$ is analogous.
%
The dual variable $\mathbf{Z}$ is defined as $(\mathbf{Z_1}, \mathbf{Z_2})$.
Such matrices are not separable in \Cref{latent-time-admm}, thus they must be jointly updated.
The update of $\mathbf{Z}$ in \eqref{eq:admm-algorithm} can be rewritten as follows:
\small
\begin{align}\label{eq:update-z}
\begin{split}
\left[\begin{array}{c}Z_{1,i}^{k+1}\\ Z_{2,i}^{k+1} \end{array}\right] &=
\underset{Z_{1},Z_{2}}{\argminop}~ \beta ~\Psi(Z_{2} - Z_{1})
\begin{aligned}[t]
&+\frac{\rho}{2}\|\Theta^k_i - Z_{1} + X^k_{1,i}\|^2\\
&+\frac{\rho}{2}\|\Theta^k_{i+1} - Z_{2} + X^k_{2,i}\|^2.\end{aligned}
\end{split}
\end{align}\normalsize
Let \small $\hat{\Psi}\bigg[\begin{array}{c}Z_1\\Z_2
\end{array}\bigg] = \Psi(Z_2 - Z_1)$.\normalsize
~Then, Problem \eqref{eq:update-z} can be solved with an unique update \cite{hallac2017network}:
\small
\begin{equation*}
\bigg[\begin{array}{c}
Z_{1,i}^{k+1}\\Z_{2,i}^{k+1}
\end{array}\bigg] = \text{prox}_{\frac{\beta}{\rho}\hat{\Psi}(\cdot)} \bigg(\bigg[
\begin{array}{c}
\Theta_i^k + U_{1,i}^k\\
\Theta_{i+1}^k + U_{2,i}^k
\end{array}
\bigg]\bigg).
\end{equation*}
\normalsize

The same holds for the $\mathbf{W}$ step. Hence, the proximal operator for the update of $W_{1,i}$ and $W_{2,i}$ becomes:
\small
\begin{equation*}
\bigg[\begin{array}{c}
W_{1,i}^{k+1}\\W_{2,i}^{k+1}
\end{array}\bigg] = \text{prox}_{\frac{\eta}{\rho}\hat{\Phi}(\cdot)} \bigg(\bigg[
\begin{array}{c}
L_i^k + U_{i,3}^k\\
L_{i+1}^k + U_{i,4}^k
\end{array}
\bigg]\bigg).
\end{equation*}\normalsize
For the particular derivation of different proximal operators, \mbox{see \citep{hallac2017network}.}

\subsection{Termination criterion}\label{sec:rho-update}
According to \cite{Boyd2010}, the algorithm is said to converge if the primal and dual residuals are sufficiently small, \ie\ if $\|r^k\|_2^2 \leq \epsilon^{\text{pri}}$ and $\|s^k\|_2^2 \leq \epsilon^{\text{dual}}$. At each iteration $k$
these values are computed as follows:
\small
\begin{align*}
\|r^k\|_2^2 =&~ \| \bm{R}^k \!- \bm{\Theta}^k \!+ \bm{L}^k\|_F^2 + \|P_1\bm{\Theta}^k \!- \bm{Z}_1^k\|_F^2 + \| P_2\bm{\Theta}^k- \bm{Z}_1^k\|_F^2
\\ & + \|P_1\bm{L}^k \!- \bm{W}_1^k\|_F^2 + \| P_2\bm{L}^k \!- \bm{W}_2^k\|_F^2\\
\|s^k\|_2^2 =&~ \rho \big( \|\bm{R}^k \!- \bm{R}^{k-1}\|_F^2 + \| \bm{Z}_1^k \!- \bm{Z}_1^{k-1}\|_F^2 + \|  \bm{Z}_2^k \!- \bm{Z}_2^{k-1}\|_F^2 \\&+ \| \bm{W}_1^k \!- \bm{W}_1^{k-1}\|_F^2 + \|\bm{W}_2^k \!- \bm{W}_2^{k-1} \|_F^2 \big)\\
\epsilon^{\text{pri}} =&~ c+\epsilon^\text{rel}~\text{max}\big(D_1 ,D_2\big)
\\
\epsilon^{\text{dual}} =&~ c+\epsilon^\text{rel}\rho\big(
\|\bm{U}_0^k\|_F^2 +  \|\bm{U}_1^k\|_F^2 +  \|\bm{U}_2^k\|_F^2 + \| \bm{U}_3^k\|_F^2 + \|\bm{U}_4^k\|_F^2 \big)
\end{align*}\normalsize
where $c = \epsilon^\text{abs}d(5T-4)^{1/2}$, $\epsilon^\text{abs}$ and $\epsilon^\text{rel}$ are arbitrary tolerance parameters, $D_1^k = \|\bm{R}^k\|_F^2 + \| \bm{Z}_1^k\|_F^2 + \| \bm{Z}_2^k\|_F^2 + \| \bm{W}_1^k\|_F^2 + \| \bm{W}_2^k\|_F^2$ and $D_2^k = \|\bm{\Theta}^k- \bm{L}^k\|_F^2 + \| P_2\bm{\Theta}^k\|_F^2 + \|  P_1\bm{\Theta}^k\|_F^2 + \| P_2\bm{L}^k\|_F^2 + \| P_1\bm{L}^k \|_F^2$.\\


\subsection{Implementation}
The minimisation algorithm is available as a Python framework\footnote{\ltgl\ is open-source. The code is available under BSD-3-Clause at \mbox{\url{https://github.com/fdtomasi/regain}}.}, fully compatible with the \emph{scikit-learn} library of machine learning algorithms, providing a straightforward and intuitive interface. The implementation relies on low-level high-performance libraries for numerical computations and it exploits closed-form solutions for proximal operators, leading to a fast and scalable minimisation algorithm even with an increasing number of unknowns.

\section{Experiments}\label{synthetic}
\relax\sloppy
We performed experiments on synthetic data assessing the performance of the method in terms of structure recovery and measure of latent variables influence.
The performance of \ltgl\ was evaluated with respect to the ground truth and to state-of-the-art methods for graphical inference. In particular, we assessed two aspects of such methods, that are \emph{modelling performance} and \emph{scalability}, in separated experiments. {Modelling performance} was estimated by comparing the inferred graphical model to the true network underlying the data set. During the scalability experiment, instead, we assessed the computational time for convergence needed for increasing problem complexity.
%
%
\subsection{Modelling performance}\label{sec:accuracy}
We evaluated \ltgl\ {modelling performance} on two synthetic data sets. 
%
%
The {ground truth} sets of matrices $\mathbf{\Theta}~=~(\Theta_1, \dots, \Theta_T)$ and  $\mathbf{L}~=~(L_1, \dots, L_T)$ were obtained by perturbing initial matrices $\Theta_1$ and $L_1$, according to a specific behaviour for $T-1$ times, with the guarantee that $\Theta_i - L_i \succ 0 $ and $L_i \succcurlyeq 0 $ for $i=1,\dots,T$.
The initial matrices
were generated according to \cite{yuan2012}, following the form \eqref{schur}. $\Theta_1$ and $L_1$ correspond to $\Theta_O$ and $\Theta_{OH}\Theta_H^{-1}\Theta_{HO}$, respectively, with $\Theta_H$ identity matrix and $\Theta_{HO}= \Theta_{OH}^\top$.
Note that, since $\Theta_H$ has full rank, the number of latent variables is $H$.
%
In particular, for $d$ observed variables, $n$ samples and $T$ timestamps, we generated a data set \mbox{$X \in (\R^{n\times d})^T$} sampled from $T$ multivariate normal distributions $X_i \sim \mathcal{N}_i(\mathbf{0}, \Sigma_{i})$, for \mbox{$i=1,\dots,T$}, and $\Sigma^{-1}_{i} = \Theta_i - L_i$.

\subsubsection{$\ell_2^2$ perturbation $(p_2)$}
The first data set was generated by perturbing the initial matrices with a random matrix of small $\ell_2^2$ norm.
This perturbation assumes the differences between two consecutive matrices to be small and bounded over time, \ie, $\|\Theta_i - \Theta_{i-1}\|_F \leq \epsilon$ for $i=2,\dots,T$.
The bound $\epsilon$ on the norm is chosen \emph{a priori}.

The update of $L_i$ is done maintaining consistency with the theoretical model where $L_i = \Theta_{OH,i}\Theta_{OH,i}^\top$. Therefore, the update is obtained by adding a random matrix with a small norm to $\Theta_{OH,i-1}$. In this way, the rank of $L_i$ remains the same as the number of latent variables and constant over time.
Data were generated in $\R^{100}$ with 10 time stamps, conditioned on 20 latent variables. For each time stamp, we drew 100 samples from the distribution. For this reason, in this setting, the contribution of latent factors is predominant with respect to the network evolution in time.

\vspace*{-.5em}
\subsubsection{$\ell_1$ perturbation $(p_1)$}
A second data set was generated according to a different perturbation model.
Here, the precision matrix was updated by randomly choosing an edge and swapping its state, \ie, by removing or adding a connection between two variables.
This allows for a $\ell_1$-norm evolutionary pattern of the network.
Data were generated in $\R^{50}$ with 100 time stamps, conditioned on 5 latent variables. For each time stamp, we drew 100 samples from the distribution.
In this setting, the time consistency affects the network more than the latent factor contribution.

\begin{table}[t]
  \centering
  \caption{\small{Performance in terms of $F_1$ score, accuracy (ACC), mean rank error (MRE) and mean squared error (MSE) of \ltgl\ with respect to TVGL, LVGLASSO and GL.
  \ltgl\ and TVGL are employed with both $\ell_2^2$ and $\ell_1$ penalties, to show how the prior on the evolution of the network affects the outcome.}
  }
  \begin{tabular}{llrrcc}
    \toprule
    \multirow{2}{*}{perturbation} & \multirow{2}{*}{method} & \multicolumn{4}{c}{score} \\
    && $F_1$ &  ACC & MRE & MSE \\
    \midrule
    \multirow{6}{*}{$\ell_2^2$ ($p_2$)} & LTGL ($\ell_2^2$) &  \textbf{0.926} &  \textbf{0.994} & \textbf{0.70} & 0.007 \\
    &LTGL ($\ell_1$)    &  0.898 &     0.993 & \textbf{0.70} & 0.007  \\
    &TVGL ($\ell_2^2$)  &  0.791 &     0.980 & -   & \textbf{0.003} \\
    &TVGL ($\ell_1$)    &  0.791 &     0.980 & -   & \textbf{0.003} \\
    &LVGLASSO           &  0.815 &     0.988 & 2.80 & 0.007 \\
    &GL                 &  0.745 &     0.974 &   - & 0.004 \\
    \midrule
    \multirow{6}{*}{$\ell_1$ ($p_1$)}
    &LTGL ($\ell_2^2$) &  0.842 & 0.974 &  0.29 & 0.013\\
    &LTGL ($\ell_1$)    & \textbf{0.880} &     \textbf{0.981} & \textbf{0.28}&0.013 \\
    &TVGL ($\ell_2^2$)  & 0.742 &     0.950 &            - &0.009\\
    &TVGL ($\ell_1$)    & 0.817 &     0.968 &            - &0.009\\
    &LVGLASSO           & 0.752 &     0.964 &          0.74 &0.013\\
    &GL                 &  0.748 &     0.951 &     - &\textbf{0.007}\\
    \bottomrule\\
  \end{tabular}
  \label{tab:results}
  \vspace*{-2em}
\end{table}
\begin{figure}[t]
  \includegraphics[width=.48\textwidth]{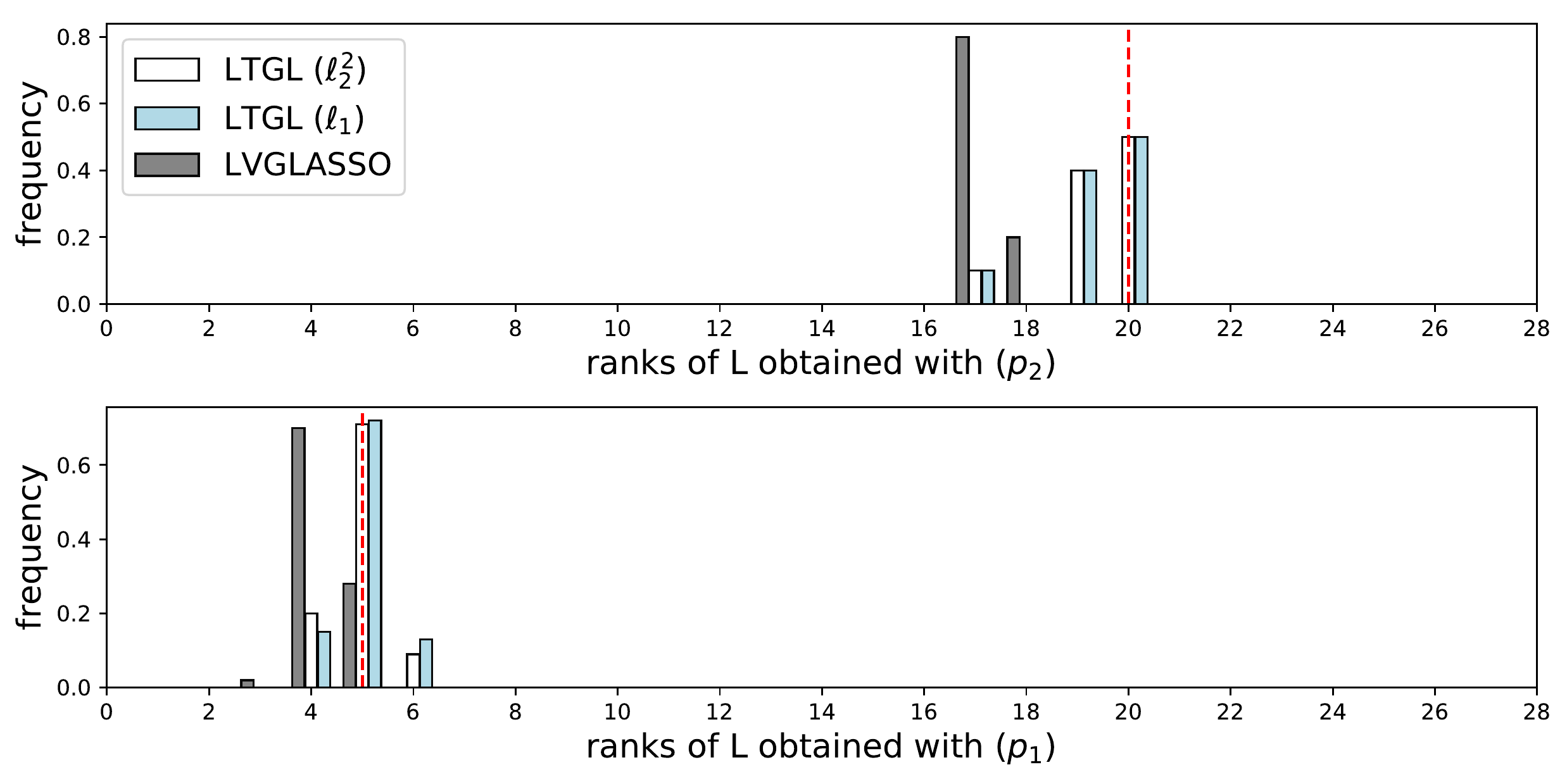}
  \vspace*{-2em}
  \caption{\small{Distribution of inferred ranks over time.
  For each method that considers latent variables, we show the frequency of finding a specific rank during the network inference.
  The vertical line indicates the ground truth rank, around which all detected ranks lie.
  Note that, in $(p_2)$, $L_i\in\R^{100\times 100}$, therefore the range of possible ranks is $[0,100]$. For $(p_2)$, $L_i\in\R^{50\times 50}$, hence the range is $[0,50]$.}}
  \label{fig:ranks}
  \vspace*{-1em}
\end{figure}
\begin{figure*}[t!]
  \includegraphics[width=.95\textwidth]{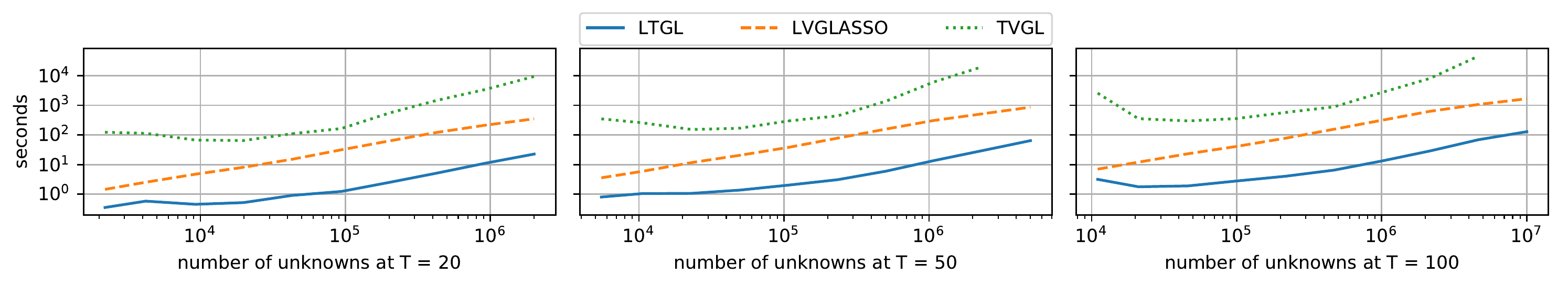}
  \vspace*{-1.5em}
  \caption{\small{Scalability comparison for \ltgl\ in relation to other ADMM-based methods. The compared methods are initialised in the same manner, \ie, with all variable interactions (not self-interacting) set to zero.
  For LVGLASSO and TVGL, we used their relative original implementations.
  Also, we ignore the computational time required for hyperparameters selection.
  \ltgl\ outperforms the other methods for each increasing time and dimensionality of the problem.}}
  \label{fig:scalability}
  \vspace*{-1em}
\end{figure*}

\subsubsection{Scores}
We evaluated \ltgl\ performance using different scores measuring the divergence of the results from the ground truth. In particular, the performance was evaluated in terms of $F_1$ score, accuracy, mean rank error and mean squared error.
We define as \emph{true/false positive} the number of correctly/incorrectly existing inferred edges, \emph{true/false negative} the number of correctly/incorrectly missing inferred edges \cite{hecker2009gene}. The scores are computed as follows.\\[.5em]
{\it\textbullet\ $F_1$ score}:
indicates the quality of structure inference,
as the harmonic mean of precision and recall. \\
{\it\textbullet\ Accuracy (ACC)}: evaluates the number of true existing and missing connections in the network correctly inferred with respect to the total number of connections. \\
{\it\textbullet\ Mean rank error (MRE)}: estimates the precision on the number of inferred latent variables, based on the rank of the set of matrices $\tilde{\mathbf{L}}$ in relation to the ground truth. The MRE score is defined as:
\small $$\text{MRE} =  \frac{1}{T} \sum_{i=1}^T \big| \text{rank}(L_i) - \text{rank}(\tilde{L}_i) \big|.$$\normalsize
A value close to 0 means that we are inferring the true number of latent variables over time, while, viceversa, a high value indicates a poor consideration of the contribution of the latent variables.\\
{\it\textbullet\ Mean squared error (MSE)}: score how close is the inferred precision matrix $\tilde{\mathbf{\Theta}}$ to the ground truth, in terms of the Frobenius norm:
\small $$\text{MSE} =  \frac{2}{Td(d-1)} \sum_{i=1}^T \norm{\Theta^{(u)}_i - \tilde{\Theta}^{(u)}_i }_{F},$$
\vspace*{-.5em}\normalsize
where $\Theta^{(u)}$ denotes the upper triangular part of $\Theta$.
%
\subsubsection{Discussion}
\Cref{tab:results} shows the performance of \ltgl\ compared to graphical lasso (GL) \cite{friedman2008sparse}, latent variable graphical lasso (LVGLASSO) \cite{chandrasekaran2010latent, Ma2013} and time-varying graphical lasso (TVGL) \cite{hallac2017network} in terms of $F_1$ score, accuracy, mean rank error (MRE) and mean squared error (MSE), for both settings with $\ell_2^2$ $(p_2)$ and $\ell_1$ $(p_1)$ perturbation.
Note that MRE is not available for all the methods since neither GL or TVGL consider latent factors.
\ltgl\ and TVGL are used with two temporal penalties according to the different perturbation models of data generation. In this way, we show how the correct choice of the penalty for the problem at hand results in a more accurate network estimation.
In both $(p_2)$ and $(p_1)$, \ltgl\ outperforms the other methods for graphical modelling. In $(p_2)$, in particular, \ltgl\ correctly infers almost 99,5\% of edges in all the dynamical network both with the $\ell_2^2$ and $\ell_1$ penalties. Nonetheless, the use of $\ell_2^2$ penalty enhance the quality of the inference as expected from the theoretical assumption made during data generation.
The choice of a proper penalty for the problem and the consideration of time consistency is reflected also in a low MRE, which encompasses LVGLASSO ability in detecting latent factors (\Cref{fig:ranks}). In $(p_2)$, in fact, the number of latent variables with respect to both observed variables and samples is high. Therefore, by exploiting temporal consistency of the network, \ltgl\ is able to improve the latent factors estimation.
%
%
Simultaneous consideration of time and latent variable also positively influences the $F_1$ score, \ie, structure detection.

Above considerations also hold for the $(p_1)$ setting. Here, \ltgl\ achieves the best results in both $F_1$ score and accuracy, while having a low MRE. The adoption of $\ell_1$ penalty improves structure estimation and latent factors detection, consistently with the data generation model.
%
%
%
Such settings were designed to show how the prevalence of latent factors contribution or time consistency affects the outcome of a network inference method.
In $(p_2)$, where the latent factors contribution is prevalent, network inference is more precise when considering latent factors. In $(p_1)$, instead, the number of time points is more relevant than the contribution of latent factors, hence it is more effective to exploit time consistency (both for latent and observed variables), evident from the results of \Cref{tab:results}.
\ltgl\ benefits from both aspects, therefore leading to a noticeable improvement of graphical modelling.
%

\vspace*{-.5em}
\subsection{Scalability}
Next, we performed a scalability analysis using \ltgl\ with respect to different ADMM-based solvers. We evaluated the performance of our method in relation to LVGLASSO \cite{Ma2013} and TVGL \cite{hallac2017network}, both implemented with closed-form solutions to ADMM subproblems.
In general, the complexity of the three compared solvers is the same (up to a constant). 
The implementation of GL \cite{friedman2008sparse} was not included in such experiment, since it is not based on ADMM but on coordinate descent, 
and therefore it is not comparable to our method. 
As in \Cref{sec:accuracy}, we generated different data sets \mbox{$X \in (\R^{n\times d})^T$} with different values of $T$ and $d$. In particular, $d \in [10, 400)$ and $T = \{20,50,100\}$.
We ran our experiments on a machine provided with two CPUs (2.4 GHz, 8 cores each).

\Cref{fig:scalability} shows, for the three different time settings, the scalability of the methods in terms of seconds per convergence considering different number of unknowns of the problem (\ie, $2T\frac{d(d+1)}{2}$ with $d$ observed variables and $T$ times).
In all settings, \ltgl\ outperforms LVGLASSO and TVGL in terms of seconds per convergence.
In particular, the computational time for convergence remains stable disregarding the number of time points under consideration.
We emphasise that
the most computationally expensive task performed by our solver is represented by
two eigenvalue decompositions, with a complexity of $O(d^3)$, to solve both $\mathbf{R}$ and $\mathbf{L}$ steps (\Cref{mm}).

\vspace*{-.5em}
\subsection{Model selection}
The hyperparameters of the method have been selected by using a cross-validation procedure.
%
%
%
In particular, we used the Monte Carlo Cross-Validation (MCCV) \cite{molinaro2005prediction} that repeatedly splits the $n$ samples of the data set in two mutually exclusive sets. For each split, $n\cdot(1/\nu)$ samples are labelled as {\em validation} set and the remaining $n\cdot(1 - 1/\nu)$ as {\sl learning} set. 
%
For each hyperparameter combination, the model was trained on the learning set and the likelihood of the model was estimated on the independent test set. 
%
Finally, we selected the combination of hyperparameters based on the average maximum likelihood of the model across multiple splits of the data set.

However, the number of possible combinations of \ltgl{} hyperparameters can be arbitrarily large.
In order to avoid the assessment of a grid of models (which can be computationally expensive), we used a Gaussian process-based Bayesian optimisation procedure to choose the best combination of hyperparameters for each analysed data set, based on the Expected Improvement (EI) strategy \cite{snoek2012practical}.
In practice, assuming the dynamics of a real system to be unknown, it is possible to select the most appropriate temporal penalty by exploiting the same principles, \ie, via a model selection procedure based on the likelihood of different temporal models.

\section{Applications to real data}\label{real}
We applied \ltgl{} to two real data sets, to show how the method can be employed to infer useful insights on multivariate time-series data. These data sets measure complex dynamical systems of different (biological and financial) nature, which are usually highly dimensional and feature complicated interdependences between variables. This fact makes them ideal candidates for an analysis using graphical models.

\vspace*{-.5em}
\subsection{Metabolomic Data}\label{sec:application-metabolomic}
The physiology of \Ecoli{} necessitates rapid changes of its cellular and molecular network to adapt to environmental conditions. \ecoli{} is widely studied because of the efficiency in its system response to perturbation.
Following the analysis of \cite{jozefczuk2010metabolomic}, we used \ltgl{} on \ecoli{} data to infer network modifications across different time points evaluated before and after the application of environmental condition perturbations. We analysed the behaviour of metabolites, which have been shown to change consistently after the perturbation. Samples underwent one of two types of stress, namely cold and heat stress.

\begin{figure}[t]
\includegraphics[width=.5\textwidth]{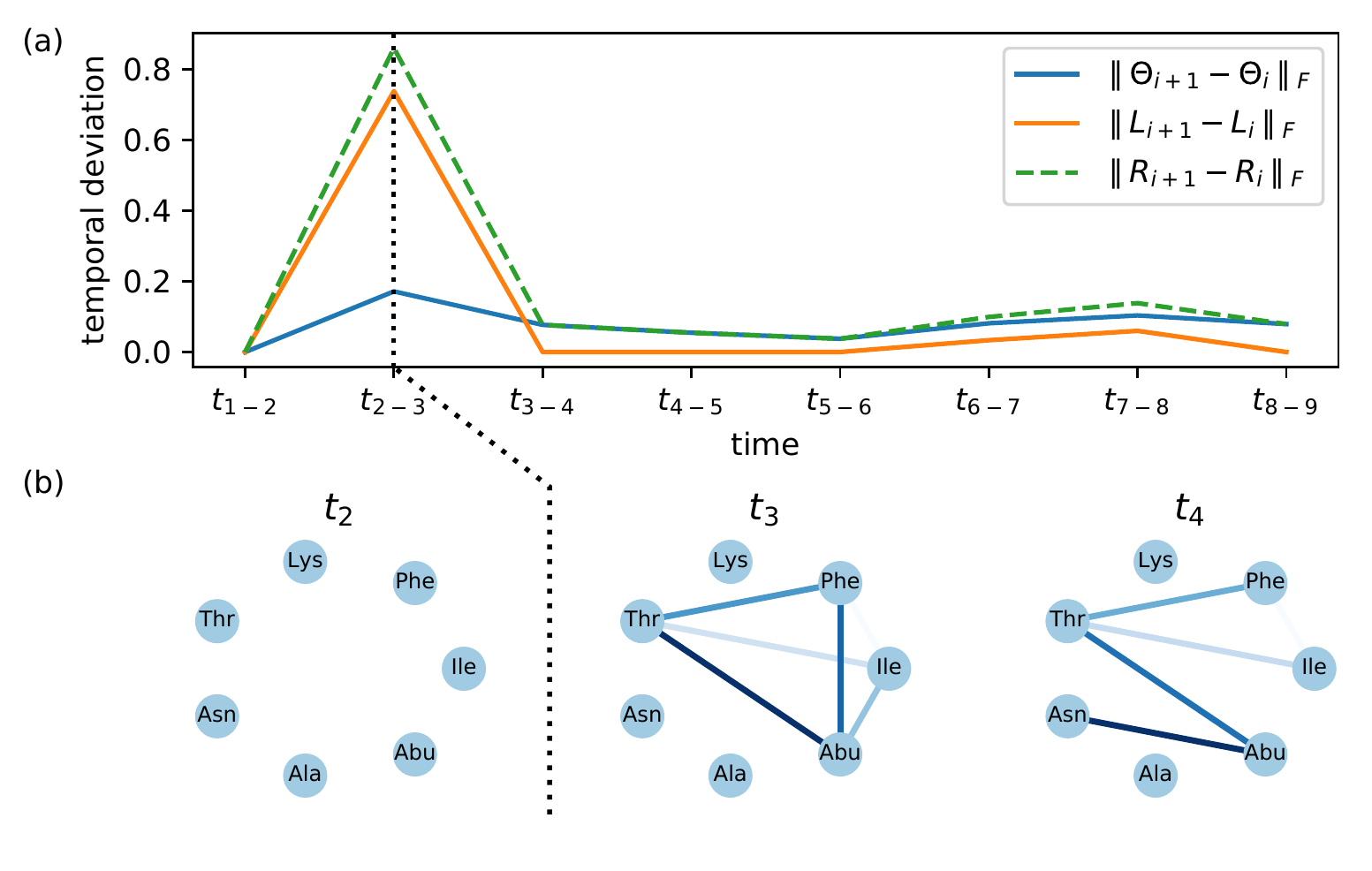}
\vspace*{-2.5em}
\caption{\small{Structure change of \ecoli{} metabolites subject to stress. The perturbation happens between time $t=2$ and $t=3$ (vertical dotted line).
(a) Temporal deviation where each point represents the difference between the network at subsequent time points. The highest deviation on the observed network $R$ appears when the stress was applied. This can be decomposed into two parts, the latent factors $L$ and the underlying structure of observed variables $\Theta$.
(b) Structural changes of metabolites interactions before and after the perturbation.}}
\label{fig:real_data_temporal_deviation}
\vspace*{-1em}
\end{figure}

\vspace*{-.5em}
\paragraph{\bf Perturbation response detection}
We inferred the dynamical network of \ecoli{} metabolites using \ltgl\ with a group lasso ($\ell_2$) penalty on latent variable contribution and a Laplacian ($\ell_2^2$) penalty on the observed network.
In this way, we allow the latent variables (which, in our model, could represent the stress or other factors) to change their global influence at a specific time point, while remaining stable in all others. At the same time, by conditioning the network on the latent variables, we allow the observed network structure to change smoothly in time.  Hence, we expect to see a global shift of the network between the second and third time points, that is when the perturbation has been introduced in the system.
\Cref{fig:real_data_temporal_deviation}a shows the temporal deviation between time points, both for $\Theta$, $L$ and the total observed system $R =  \Theta - L$.
Latent variables temporal deviation reaches a peak at time $t_{2-3}$, right after the application of the perturbation to the system. Instead, the difference between consecutive $\Theta$s remains more stable. Consistently, the difference between the observed networks $R$s shows a major change at the same time point. Hence we can distinguish the underlying evolving structure of metabolites while detecting the contribution of the latent variables which affect mostly the total system.
%
In accordance with \cite{jozefczuk2010metabolomic}, we observed a interaction
between isoleucine, threonine, phenylalanine and 2-aminobutyric acid
during the adaptation phase following the stress response (\Cref{fig:real_data_temporal_deviation}b).
Therefore, we can conclude that \ltgl\ successfully inferred a dynamical network which adjusts in response to perturbation, in accordance with our prior knowledge about \ecoli{} behaviour.

\subsection{Stock market}
\begin{figure}[t]
\includegraphics[width=.5\textwidth]{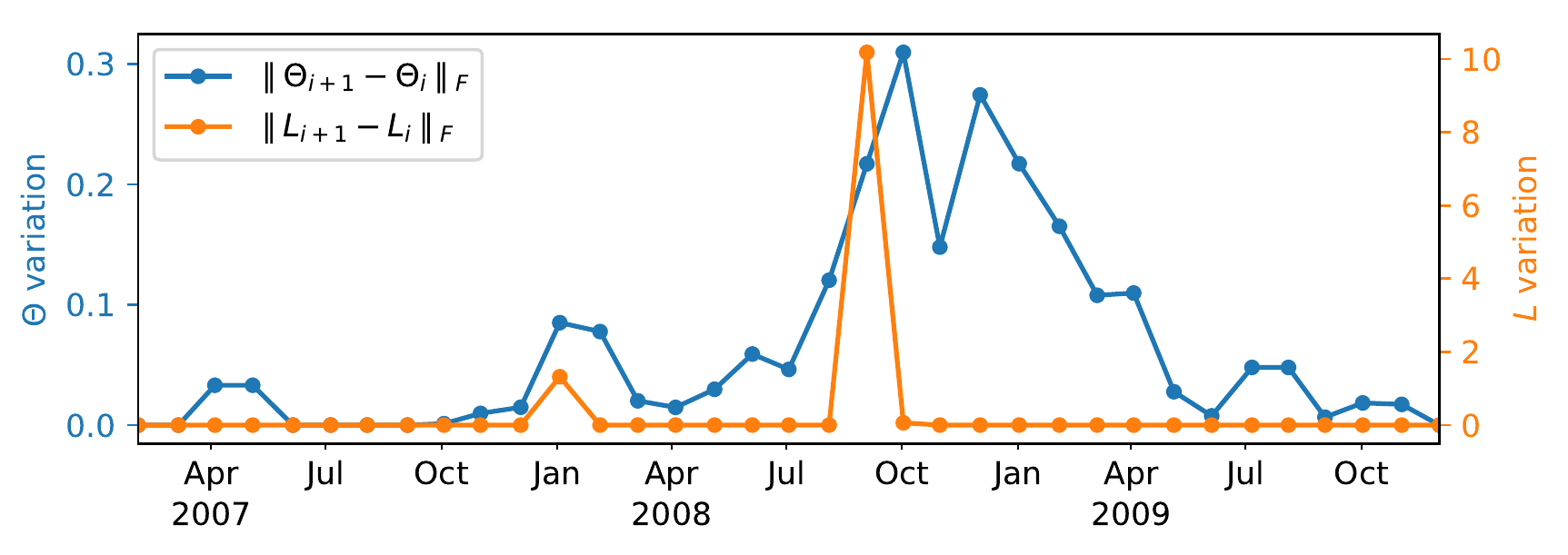}
\vspace*{-2em}
\caption{\small{Temporal deviation for stock market data. We observed two peaks, in correspondence of late 2007 and late 2008, when the financial crisis happened.}}
\label{fig:global_shift_financial_crisis}
\vspace*{-1em}
\end{figure}

Finance is another example of a complex dynamical system suitable to be analysed through a graphical model.
Stock prices, in particular, are highly related to each other and subject to time and environmental changes, \ie, events that modify the system behaviour but are not directly related to companies share values \cite{bai2006evaluating}.
%
Here, the assumption is that each company, while being part of a global financial system, is directly dependent from only a subset of others.
For example, it is reasonable to expect that stock prices of a technology company are not directly influenced by trend of companies on the primary sector.
%
%
The modelling power of \ltgl\ allows to detect both the evolution of relations between companies and environmental changes happening at a particular point in time.
In order to show this, we analysed stock prices\footnote{Data are freely available on https://quantquote.com/historical-stock-data.} during the financial crisis of 2007-2008.
The experiment was designed to consider the latent influence of the market drop on technology companies interactions.
%
%
%
\paragraph{\bf Global market crisis detection}
We used a group lasso ($\ell_2$) penalty to detect global shifts of the network.
\Cref{fig:global_shift_financial_crisis} shows two major changes in both components of the network (latent and observed), in correspondence of late 2007 and late 2008.
In particular, during October 2008 a global crisis of the market occurred, and this effect is especially evident for the shift of latent variables.
Also, the observed network changes in correspondence of the latent variables shift or immediately after, caused by the effect of the crisis on the stock market.
The latent factors influence explains how the change of the network was due to external factors that globally affected the market, and not to normal evolution of companies relationships.
We further investigated on the causes for the first shift. Indeed, we found that in late 2007 it happened a drop of a big American company that was later pointed out as the beginning of the global crisis of the following year.
%

\section{Conclusions and future work}\label{sec:conclusions}
In this work, we developed a novel method for graphical modelling of multivariate time-series.
The model considers simultaneously the contribution of latent factors and time consistency in evolving systems.
Indeed, our work is an attempt to generalise both latent variable and dynamical network inference.
To this aim, we impose prior knowledge on the problem through penalty terms that force precision and latent matrices to be consistent in time.
The choice of proper penalty terms maintains the convexity of the minimised functional and, along with the coercivity given by the regularisers, it guarantees global convergence of the proposed minimisation algorithm.
Our experiments demonstrate the ability of \ltgl\ in the graphical modelling of synthetic and real-world data, 
where the possibility to decompose the total network into two separated components allows for a better understanding of the underlying phenomenon.

We emphasise that our framework is modular in the choice of the penalties, allowing for precise modelling of different and complex behaviours of the system.
This allows for a straightforward inclusion of additional penalty terms, based on the prior knowledge on the problem at hand.
Possible extensions may involve alternative evolutionary models for different complex systems, 
\eg\, forcing subgroups of variables to behave consistently in time \cite{bolstad2011causal}.
These could lead to interesting results in time-series clustering and pattern discovery.
Further investigations may also head to the inference of the exact contribution of latent factors starting from the $\mathbf{L}$ matrices we are estimating, possibly using matrix factorisation methods \cite{ding2005equivalence}.
Such developments may increase the expression power of the method, leading to advances in data mining and to potential applications in diverse science fields.
\vspace{1em}
  \textbf{Acknowledgments.}
  The authors thank the anonymous reviewers for their valuable comments.
%
%
%

\bibliographystyle{ACM-Reference-Format}
\bibliography{bibliography}

\end{document}